\documentclass[conference]{IEEEtran}
\IEEEoverridecommandlockouts
\usepackage{multirow}
\usepackage{cite}
\usepackage{amsmath,amssymb,amsfonts}
\usepackage{algorithmic}
\usepackage{graphicx}
\usepackage{textcomp}
\usepackage{xcolor}
\usepackage{float}
\usepackage[font=footnotesize]{subfig}
\usepackage{mathtools}
\usepackage{array}
\usepackage{multirow}
\usepackage{url} 
\usepackage{graphicx}
\usepackage{booktabs}
\usepackage{adjustbox,lipsum}
\usepackage{stfloats}
\def\BibTeX{{\rm B\kern-.05em{\sc i\kern-.025em b}\kern-.08em
    T\kern-.1667em\lower.7ex\hbox{E}\kern-.125emX}}
\begin{document}

\title{Enabling Next-Generation Consumer Experience with Feature Coding for Machines}
\author{%
Md Eimran Hossain Eimon, Juan Merlos, Ashan Perera, \\ Hari Kalva, Velibor Adzic, and Borko Furht\\[0.5em]
{\small
\begin{minipage}{\linewidth}
\begin{center}
Florida Atlantic University \\
\url{{meimon2021, jmerlosjr2017, aperera2016, hkalva, vadzic, bfurht}@fau.edu} 
\end{center}
\end{minipage}
}
}
% \IEEEauthorblockA{\textit{Department of Electrical Engineering and Computer Science} \\
% \textit{Florida Atlantic University}\\
% \and
% \IEEEauthorblockA{\textit{Department of Electrical Engineering and Computer Science} \\
% \textit{Florida Atlantic University}\\
% \and
% \IEEEauthorblockA{\textit{Department of Electrical Engineering and Computer Science} \\
% \textit{Florida Atlantic University}\\
% \and
% \IEEEauthorblockA{\textit{Department of Electrical Engineering and Computer Science} \\
% \textit{Florida Atlantic University}\\
% \and
% \IEEEauthorblockA{\textit{Department of Electrical Engineering and Computer Science} \\
% \textit{Florida Atlantic University}\\
% \and
% \IEEEauthorblockA{\textit{Department of Electrical Engineering and Computer Science} \\
% \textit{Florida Atlantic University}\\

\maketitle

\begin{abstract}
As consumer devices become increasingly intelligent and interconnected, efficient data transfer solutions for machine tasks have become essential. This paper presents an overview of the latest Feature Coding for Machines (FCM) standard, part of MPEG-AI and developed by the Moving Picture Experts Group (MPEG). FCM supports AI-driven applications by enabling the efficient extraction, compression, and transmission of intermediate neural network features. By offloading computationally intensive operations to base servers with high computing resources, FCM allows low-powered devices to leverage large deep learning models. Experimental results indicate that the FCM standard maintains the same level of accuracy while reducing bitrate requirements by 75.90\% compared to remote inference.
\end{abstract}


\section{Introduction}
\begin{figure}[t]
    \centering
    \begin{minipage}[b]{1\linewidth}
    \centering
    \includegraphics[width=\textwidth]{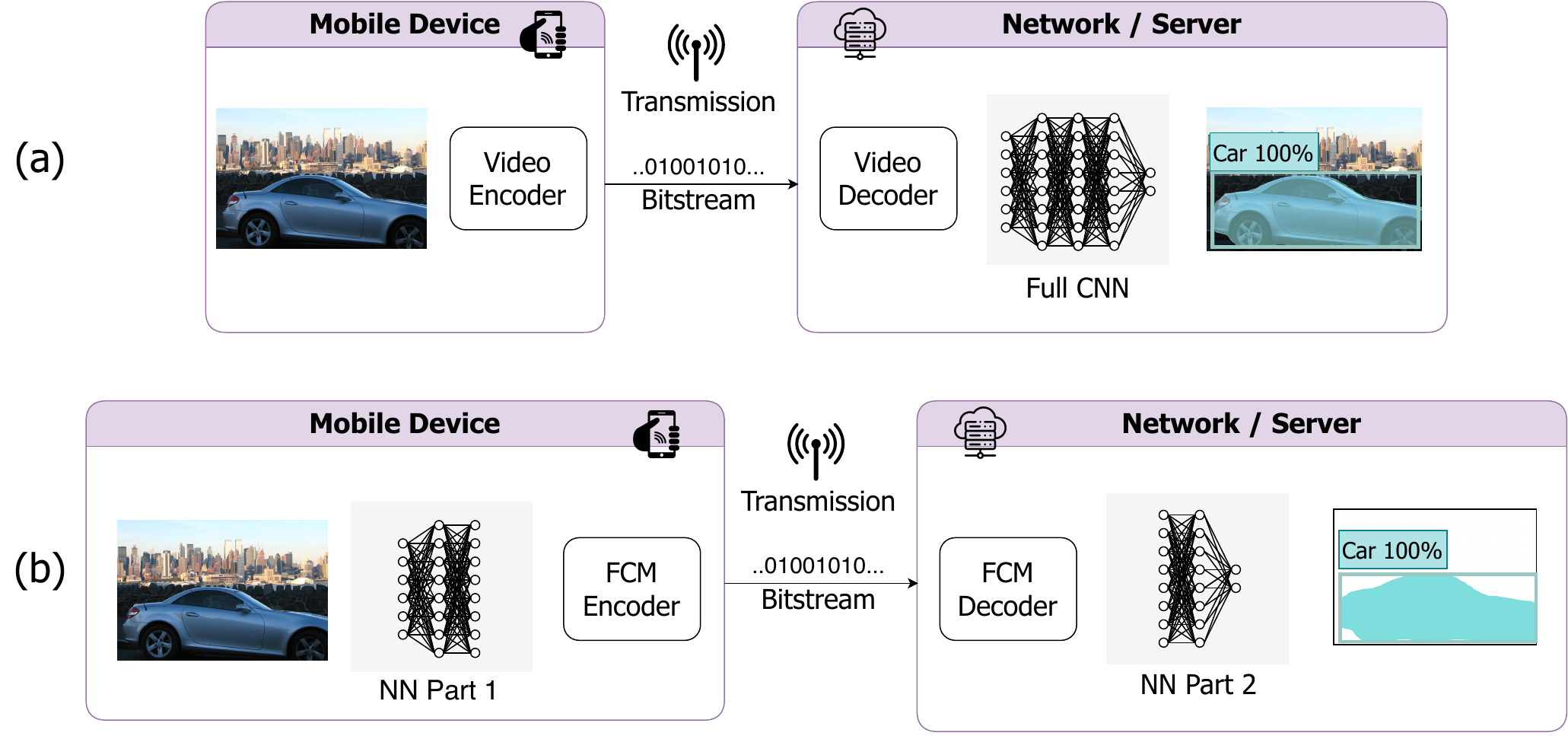}
    %\centerline{}\medskip
    \end{minipage}
\caption{(a) Remote Inference (b) Split Inference}
\vspace{-0.5cm}
\label{fig:vcm_fcm}
\end{figure}

As consumer devices continue to evolve, becoming more intelligent and interconnected, machines have increasingly become the consumer of information. A growing need emerges for solutions that effectively handle the transmission of information for neural network-based tasks. Remote inference, as shown in Fig.~\ref{fig:vcm_fcm}(a), is one such solution, where data is collected at a device level but processed remotely. This method enables consumer electronics like smartphones, augmented reality (AR) glasses, and other IoT devices to leverage powerful deep learning models that they otherwise could not execute locally. In this setup, a device may capture video, but a server processes the neural network calculations, sending results back to the device. However, remote inference, while useful, can face limitations. For instance, it may not always scale well for large networks of devices, as the data load can overwhelm both the network and the server. In light of these challenges, more advanced methodologies like split inference (a.k.a. collaborative intelligence \cite{colabintel}) have gained attention as a more efficient alternative.

% \begin{figure}[b]
%     \centering
%     \includegraphics[width=1\linewidth]{images/fig_1_b.png}
%     \caption{Remote Inference}
%     \vspace{-0.5cm}
%     \label{fig:fcm_simple}
% \end{figure}

Split inference refers to the division of a neural network into two parts: the head (NN Part 1) and the tail (NN Part 2). The head of the network is executed on the local device, while the tail is executed on a remote server. The intermediate feature data produced by the first part of the network is sent from the device to the server, where the remaining computations are completed. This collaborative approach is more efficient than transmitting raw data to a server, as the feature data can be made into a more compact and meaningful representation of the input for the neural network. An overview of this process can be seen in Fig.~\ref{fig:vcm_fcm}(b).

% \begin{figure}[b]
%     \centering
%     \includegraphics[width=1\linewidth]{images/fig_1_c.png}
%     \caption{Split Inference}
%     \vspace{-0.5cm}
%     \label{fig:fcm_simple}
% \end{figure}

The significance of split inference lies in its ability to balance computational load and optimize energy use for devices with limited resources. By offloading only part of the deep learning model's computation to a remote server, split inference reduces the amount of data that needs to be transmitted over the network. This not only conserves bandwidth but also speeds up the overall process by allowing the device to perform some of the computations locally. Moreover, split inference, in principle, contributes to energy efficiency, prolonging battery life, and ensuring that even low-powered devices can participate in sophisticated AI-driven tasks.

Recognizing the importance of split inference, the Moving Picture Experts Group (MPEG) established an Ad-Hoc Group in July 2019 to start the standardization efforts focused on Feature Coding for Machines (FCM)\cite{fcm_cttc} to support split inference in various practical scenarios such as smart city, smart traffic, video surveillance, consumer electronics. At October 2023, MPEG Working Group 4 (WG4) began FCM by evaluating 12 baseline proposals from different organizations, including Canon, InterDigital, ETRI, Sharp, Digital Insights Inc., China Telecom, ZJU, KHU, HNU, KAU, FAU~\cite{press_release_mpeg144}. 

FCM focuses on the compression and transmission of intermediate neural network features. Since these features can be larger in volume than the raw input data, efficient compression techniques are essential to ensure that the data can be sent over the network without overwhelming available bandwidth. FCM aims to enable resource-limited devices to efficiently transmit feature data for further processing on more powerful remote servers. 

The rest of the paper is organized as follows: Section \ref{sec:tools} briefly describes the tools of FCM, Section \ref{sec:exp_results} provides experimental results in terms of BD-rate performance and complexity analysis, and Section \ref{sec:conclusion} concludes the paper.

\section{Tools Description of FCM}
\label{sec:tools}
\begin{figure*}[t]
    \centering
    \begin{minipage}[b]{1\linewidth}
    \centering
    \includegraphics[trim={1.5cm 0 0 0}, clip, width=0.8\textwidth]{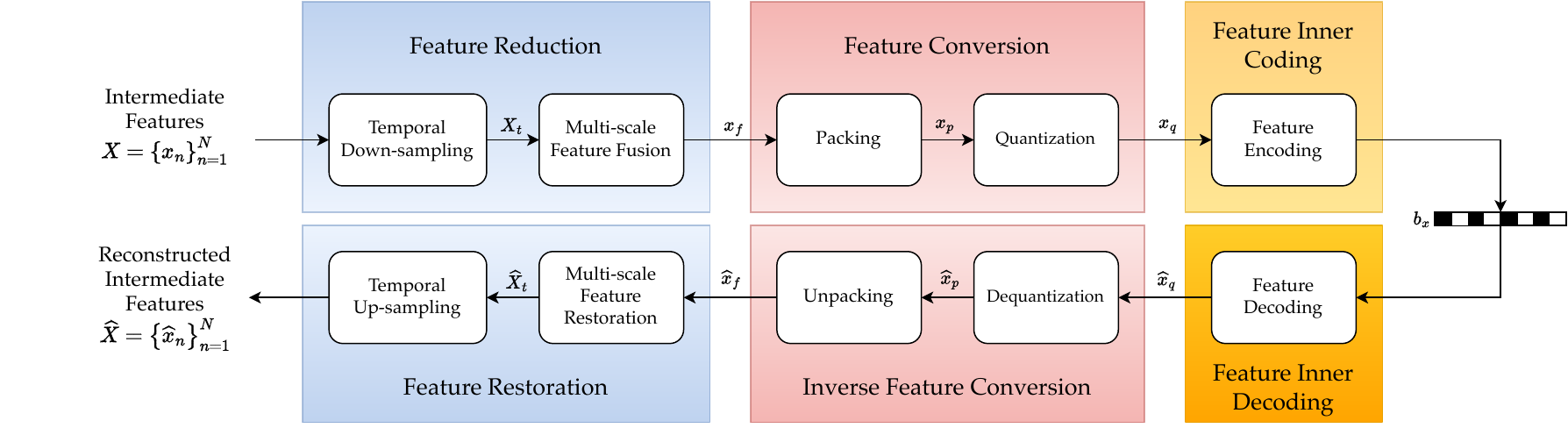}
    %\centerline{}\medskip
    \end{minipage}
\caption{Overview of Feature Coding for Machines (FCM)}
\label{fig:fcm_overview}
\end{figure*}

The following section overviews the tooling description of FCM as used for result generation and as denoted in the official algorithm description for the reference implementation, FCTM (Feature Coding Test Model) Version 3 \cite{descFCTM}. 
As depicted in Fig.~\ref{fig:fcm_overview}, encoder tools of FCM can be categorized into the following transformations on feature data: Feature Reduction, Feature Conversion, and Feature Inner coding. Decoder tools likewise are categorized into inverse transformations: Feature Inner Decoding, Inverse Feature Conversion, and Feature Restoration. 
\subsection{Feature Reduction}
\subsubsection{Temporal Down-sampling}
Temporal redundancies may exist when feature frames have little motion. Instead of transmitting every frame, FCM can temporally sample frames \(X\) based on a fixed interval, producing frames \(X_t\). A sampling ratio of 2x is supported, meaning every other frame is dropped on the encoder. Inner encoder configuration settings are changed to reflect a Group of Pictures (GOP) with size 8 shrunken to 4. Temporal resampling is turned off by default, in which case \(X_t = X\). 

\subsubsection{Multi-scale Feature Fusion}
FCM performs feature reduction using neural networks. The Feature Fusion and Encoding Network (FENet) fuses multiple feature layers \(X_t\) into a single smaller one \(x_f\), ideal for compressing with the inner codec. This is done using a series of connected \textit{Encoding Blocks} consisting of convolutional layers, residual blocks, and attention blocks. Each encoding block spatially downscales the feature layer by 2x. The feature layers are sorted by order and each is expected to be half the size of their previous one, so that the output of the first Encoding Block can be concatenated to the second feature layer before passing to the second Encoding Block. This process repeats until all feature layers have been passed through the Encoding Block(s). As different split points may have a different number of feature layers, it may not be possible to reuse the same FENet architecture across all split points. The last output from the Encoding Blocks series represents the fused features, which is spatially half the size of the smallest feature layer. As a final step, the fused features are multiplied against a gain vector used to control channel quality, producing \(x_f\). The FENet is responsible for the majority of rate-distortion improvement during feature reduction in FCM.

\subsection{Feature Conversion}
\subsubsection{Packing}
After feature reduction, the number of channels in the \(x_f\) feature tensor can be greater than the maximum supported channels by the inner codec. To address this, feature maps are spatially rearranged into a single channel frame, following a raster scan order. The number of feature maps per row and per column are determined to be as close as possible so the frame is square-like. The final feature frame \(x_p\) is suitable for any inner codec that supports monochrome coding. The original shape of \(x_f\) is also signalled to the decoder to undo packing. 

\subsubsection{Quantization}
In this step, a 10-bit linear quantization is applied. The minimum and maximum values of \(x_p\) are coded into the bitstream as floating point 32-bit for the inverse quantization at the decoder side. Feature values are quantized using the following equations:
\begin{equation}
\textit{max\_num\_bits} = 2^{\textit{bitdepth}} - 1
\end{equation}
\begin{equation}
x_q = \left\lfloor \max\left(\min\left(\frac{x_p - x_{p,\text{min}}}{x_{p,\text{max}} - x_{p,\text{min}}}, 1\right), 0\right) \times \textit{max\_num\_bits} \right\rfloor
\end{equation}

\begin{figure}[t]
    \centering
    \begin{minipage}[b]{1\linewidth}
    \centering
    \includegraphics[width=1\textwidth]{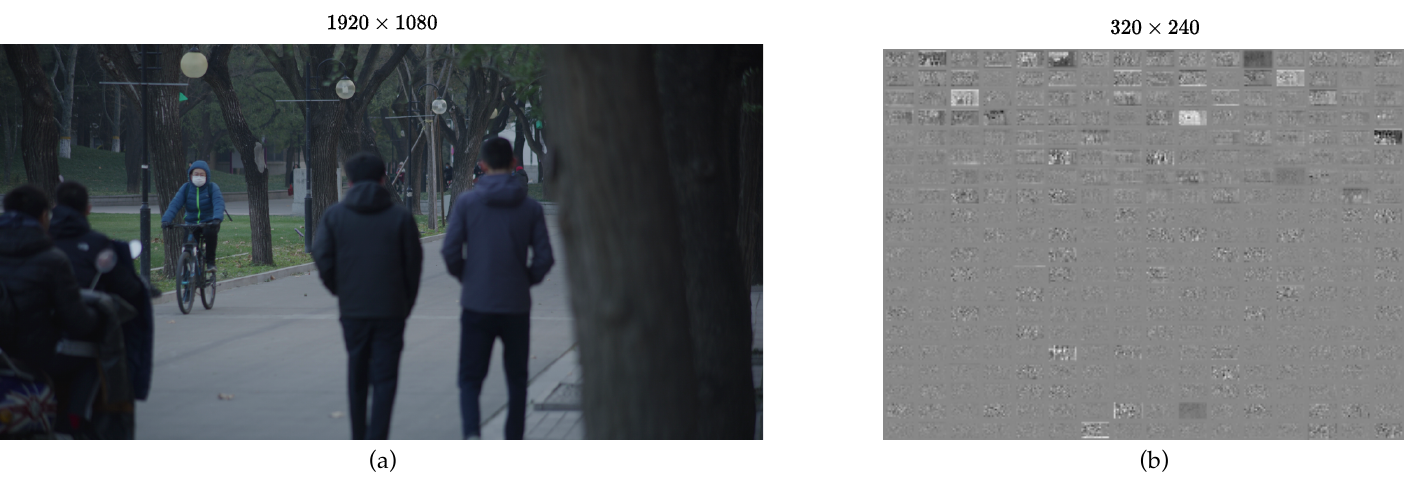}
    %\centerline{}\medskip
    \end{minipage}
\caption{(a) Input Image (b) Corresponding packed fused feature maps in YUV 4:0:0}
\vspace{-0.2cm}
\label{fig:fcm_architecture}
\end{figure}

\subsection{Feature Encoding}
In the feature encoding process, the packed feature frame as denoted in Fig.~\ref{fig:fcm_architecture}, formatted as 10-bit monochrome YUV (4:0:0), undergoes compression utilizing the Versatile Video Coding (VVC)~\cite{vvc} standard. Specifically, the encoding is performed using the VVC Test Model (VTM) version 12.0 \cite{vtm}, which provides near state-of-the-art video compression efficiency. The encoding process is configured for low-delay scenarios to simulate real-time applications. Side channel information signalled from other parts of the FCM encoder is muxed with the VVC bitstream for proper reconstruction.

\subsection{Feature Decoding}
Using the VVC VTM 12.0 decoder, the 10-bit monochrome YUV frames from the bitstream are retrieved, as well as any necessary side channel information.

\subsection{Inverse Feature Conversion}
\subsubsection{Dequantization}

The dequantization of the feature frame \( \hat{x}_p \) is computed as follows:
\begin{equation}
\hat{x}_p = \frac{\hat{x}_q}{\textit{max\_num\_bits}} \times (x_{p,\text{max}} - x_{p,\text{min}}) + x_{p,\text{min}}
\end{equation}
Here, \( \hat{x}_q \) is the reconstructed quantized feature frame, and \( x_{p,\text{min}} \) and \( x_{p,\text{max}} \) are the minimum and maximum values used for dequantization.

\subsubsection{Unpacking}

Unpacking is the reverse process of packing during feature conversion. Taking feature frame \(\widehat{x}_p\), it restores the single channel frame into multiple channels with the shape of \(x_f\). The new feature tensor is \(\widehat{x}_f\).

\subsection{Feature Restoration}
\subsubsection{Multi-scale Feature Restoration}

Feature restoration is performed on the reconstructed fused feature tensors \(\widehat{x}_f\) using neural networks. The Feature Decoding and Reconstruction Network (DRNet) is designed to perform the inverse of the encoder's FENet, by expanding \(\widehat{x}_f\) into multiple reconstructed feature layers. The fused features are first multiplied by an inverse gain vector to undo the gain operation from FENet. DRNet then splits into separate branches for each reconstructed feature layer. Each branch starts with a \textit{Decoding Block} comprised of residual blocks, attention blocks, and transposed convolutions. These will perform upscaling to restore each feature layer to their original shape in \(X_t\). Depending on the total upscaling factor, each Decoding Block will have a different repetition of layers. Besides the branch with the largest layer, each branch ends by taking the output of the Decoding Block and combining it with the previous (larger) reconstructed feature layer through a Feature Mixing Block, done to enhance the quality. The set of reconstructed feature layers from all branches is represented by \(\widehat{X}_t\). 

\subsubsection{Temporal Up-sampling}

When temporal resampling is turned on, missing frames in \(\widehat{X}_t\) are reconstructed using their two available neighbors as references. This is done using trilinear interpolation, where two of the dimensions are spatial and the third is temporal. Temporal upsampling is the last step in FCM to produce the final reconstructed intermediate features \(\widehat{X}\), ready to pass to the NN Part 2 for split inference.  

% \subsection{Reconstruction Refinement}

% Lossy compression of features coming from all stages of FCM can undesirably alter their distributions and negatively impact machine performance. FCTM attempts to readjust the distributions of restored features on the decoder by signalling feature statistics gathered on the encoder. Assuming a feature tensor [X] of length [N] follows a normal distribution, its distribution can be represented by its mean and standard deviation as described in [TODO Equations]. A reconstructed feature tensor [X'] of length [N] can be aligned to the distribution of [X] using [TODO Equation], assuming [mean X] and [sigma X] are known. Applying this process per each feature layer on the decoder corrects their distribution. Feature statistics are computed on the encoder, coded to 16-bit, and signalled to the decoder periodically. FCTM applies refinement on the restored feature maps and decoded fused feature maps separately, using the input feature maps and fused feature maps respectively for determining feature statistics.

\section{Experimental Results}
\label{sec:exp_results}
\subsection{Experimental Setup}
\begin{table}[t]
\centering
\caption{Common Test \& Training Conditions (CTTC)}
\label{tbl:cttc}
\resizebox{1\linewidth}{!}{%
\begin{tabular}{llll}
\toprule
\textbf{Dataset} & \textbf{Task} & \textbf{Network} \\ \hline 

OpenImagesV6~\cite{oiv6_seg} & Instance segmentation & MaskRCNN-X101-FPN~\cite{mask_rcnn} \\ \hline

OpenImagesV6~\cite{oiv6_det} & Object detection & FasterRCNN-X101-FPN~\cite{faster_rcnn} \\ \hline

SFU~\cite{sfu_v1} & Object detection & FasterRCNN-X101-FPN~\cite{faster_rcnn} \\ \hline

TVD~\cite{tvd} & Object tracking & JDE-1088x608~\cite{tracking} \\ \hline

HiEve~\cite{hieve} & Object tracking & JDE-1088x608~\cite{tracking} \\ \bottomrule
\end{tabular}%
}
\end{table}
We have used the MPEG Common Test \& Training Conditions (CTTC)~\cite{fcm_cttc} as the basis for our experimental framework. We have used five different dataset for three network architectures for three distinct tasks, as outlined in Table~\ref{tbl:cttc}. Encoding and decoding was done by the Versatile Video Coding (VVC)~\cite{vvc} test model VTM 12.0~\cite{vtm}. All-Intra configuration was used for encoding/decoding the image datasets, and low-delay configuration was used for encoding/decoding the video datasets. The split points for FasterRCNN-X101-FPN and MaskRCNN-X101-FPN models are defined at \(\{ p_{2}, p_{3}, p_{4}, p_{5} \}\) and the split points for JDE-1088x608 network for TVD and HiEve datasets are defined at \(\{ d_{36}, d_{61}, d_{74} \}\) and \(\{ d_{105}, d_{90}, d_{75} \}\) respectively. Three pre-defined split points have been used in the experiments to show the generalization ability of FCM. 
% \begin{equation}
% \label{eq:detectron_feat_size}
%     x_n = 
%     \begin{cases} 
%         \quad p_2 \in \mathbb{R}^{256 \times \frac{H}{4} \times \frac{W}{4}}, & \text{if } n = 1\\
%         \quad p_3 \in \mathbb{R}^{256 \times \frac{H}{8} \times \frac{W}{8}}, & \text{if } n = 2\\
%         \quad p_4 \in \mathbb{R}^{256 \times \frac{H}{16} \times \frac{W}{16}}, & \text{if } n = 3\\
%         \quad p_5 \in \mathbb{R}^{256 \times \frac{H}{32} \times \frac{W}{32}}, & \text{if } n = 4
%     \end{cases}
% \end{equation}
% \begin{equation}
% \label{eq:jde_dn_feat_size}
%     x_n = 
%     \begin{cases} 
%         \quad d_{36} \in \mathbb{R}^{256 \times 76 \times 136}, & \text{if } n = 1\\
%         \quad d_{61} \in \mathbb{R}^{512 \times 38 \times 68}, & \text{if } n = 2\\
%         \quad d_{74} \in \mathbb{R}^{1024 \times 19 \times 34}, & \text{if } n = 3
%     \end{cases}
% \end{equation}
% \begin{equation}
% \label{eq:jde_alt1_feat_size}
%     x_n = 
%     \begin{cases} 
%         \quad d_{105} \in \mathbb{R}^{128 \times 76 \times 136}, & \text{if } n = 1\\
%         \quad d_{90} \in \mathbb{R}^{256 \times 38 \times 68}, & \text{if } n = 2\\
%         \quad d_{75} \in \mathbb{R}^{512 \times 19 \times 34}, & \text{if } n = 3
%     \end{cases}
% \end{equation}

\subsection{BD-Rate Performance}
Table~\ref{tbl:bd_rate} presents the BD-Rate~\cite{bd_rate} performance of MPEG FCM in comparison to remote inference. The evaluation was conducted using CompressAI-Vision~\cite{compressai_vision} to compute model accuracy across various rate points. Employing FCM resulted in an overall BD-rate reduction of 75.90\% across all tasks and datasets, compared to remote inference. This substantial reduction in bitrate, while maintaining same accuracy for machine tasks, highlights the potential of FCM to significantly improve the efficiency of consumer electronics in upcoming decades.

\begin{table}[t]
\centering
\caption{BD-Rate Performance}
\resizebox{1\linewidth}{!}{%
\begin{tabular}{@{}l l >{\centering\arraybackslash}m{2.5cm}@{}}
\toprule
\textbf{Task Network} & \textbf{Dataset} & \textbf{FCM vs Remote Inference} \\ \midrule
\multirow{1}{*}{Instance Segmentation}  
                      & OpenImageV6      & -94.04\%                        \\ \midrule
\multirow{4}{*}{Object Detection} 
                      & OpenImageV6      & -92.37\%                        \\ 
                      & SFU (Class A/B)  & -76.64\%                        \\ 
                      & SFU (Class C)    & -80.64\%                        \\ 
                      & SFU (Class D)    & -58.77\%                        \\ \midrule
\multirow{3}{*}{Object Tracking}  
                      & TVD              & -88.50\%                        \\ 
                      & HIEVE (1080p)    & -59.91\%                        \\ 
                      & HIEVE (720p)     & -56.35\%                        \\ \midrule
                      & \textbf{OVERALL} & \textbf{-75.90\%}               \\ \bottomrule
\end{tabular}
}
\label{tbl:bd_rate}
\vspace{-0.2cm}
\end{table}

\subsection{Complexity Analysis}

Since compute offload is the main feature of FCM, it's imperative to address the complexity aspect of the FCM along with bitrate reduction. The complexity of FCM system could be analyzed using the following two equations:
\begin{equation}
\small
 \frac{\text{FCM Encoder Complexity}}{\text{NN Part 2 Complexity}} < 1
\label{eq:encoder_complexity}
\end{equation}

\begin{equation}
\small
\frac{\text{FCM Decoder Complexity}}{\text{NN Part 1 Complexity}} < 1
\label{eq:decoder_complexity}
\end{equation}
If the inequality in Eq.~\ref{eq:encoder_complexity} holds true, it indicates that FCM is computationally more efficient than edge computing as it will require less resources in edge devices. Conversely, if the inequality does not hold, it implies that the computational cost of running the entire CNN is lower than that of executing the FCM encoder. Similarly, we could argue that if the inequality in Eq.~\ref{eq:decoder_complexity} holds true, split inference is more effective than remote inference. If the inequality does not hold, it suggests that the computational cost of running the FCM decoder on the remote server exceeds that of running the full CNN. 

The complexity ratios using Eq.~\ref{eq:encoder_complexity} \& Eq.~\ref{eq:decoder_complexity} for FCM are presented in Table~\ref{tbl:complexity_analysis}. To generate the results, we have used execution time as a complexity measure and executed NN Part 1, NN Part 2, FCM Encoder \& FCM Decoder using only CPU. We have used an AMD EPYC 7702 64-Core processor to run all our experiments. 

It can be observed from Table~\ref{tbl:complexity_analysis} that the overall complexity ratio for $\frac{\text{FCM Encoder Complexity}}{\text{NN Part 2 Complexity}}$ is 11.86 which indicates that running the FCM encoder is approximately 12 times more computationally intensive than running the full CNN at the edge devices. This increased computational cost can be attributed to the FENet and DRNet modules in the feature reduction and feature restoration processes. However, for the TVD and HiEve datasets in the tracking scenario, it is possible to reduce complexity using FCM.

Furthermore, it should also be noted from Table \ref{tbl:complexity_analysis} that the overall complexity ratio for $\frac{\text{FCM Decoder Complexity}}{\text{NN Part 1 Complexity}}$ is 0.31, which implies that running the FCM decoder on the remote server requires less computational resources than running the full CNN on the remote server.

\begin{table}[t]
\centering
\caption{FCM Complexity Ratio}
\resizebox{\linewidth}{!}{%
\begin{tabular}{@{}ll>{\centering\arraybackslash}m{2.2cm}>{\centering\arraybackslash}m{2.5cm}@{}}
\toprule
\textbf{Task Network} & \textbf{Dataset} & \textbf{$\frac{\text{FCM Encoder Complexity}}{\text{NN Part 2 Complexity}}$} & \textbf{$\frac{\text{FCM Decoder Complexity}}{\text{NN Part 1 Complexity}}$} \\ \midrule
\multirow{1}{*}{Instance Segmentation}  
                      & OpenImageV6      & 15.00                              & 0.25                \\ \midrule
\multirow{4}{*}{Object Detection} 
                      & OpenImageV6      & 39.28                              & 0.29                \\ 
                      & SFU (Class A/B)  & 3.72                               & 0.34                \\ 
                      & SFU (Class C)    & 6.36                               & 0.39                \\ 
                      & SFU (Class D)    & 5.90                               & 0.43                \\ \midrule
\multirow{3}{*}{Object Tracking}  
                      & TVD              & 0.93                               & 0.14                \\ 
                      & HIEVE (1080p)    & 1.01                               & 0.06                \\ 
                      & HIEVE (720p)     & 0.97                               & 0.05                \\ \midrule
                      & \textbf{Overall} & \textbf{11.86}                     & \textbf{0.31}       \\ \bottomrule
\end{tabular}
}
\label{tbl:complexity_analysis}
\vspace{-0.2cm}
\end{table}

\section{Conclusion}
\label{sec:conclusion}
In the current era of artificial intelligence, FCM provides a means to run highly complex deep learning models on resource-constrained devices by splitting the network. Experimental results have demonstrated that significant bitrate reduction can be achieved using FCM without losing end-task accuracy. This makes the FCM system attractive for different consumer electronics. However, in principle, FCM should also facilitate computational offloading by executing portions of the split neural network on a base server with high computing resources. This is only possible if the computing power required to run the FCM codec does not exceed that of executing the full network on edge devices. Our experiments indicate that running the FCM codec is more costly than running the complete Convolutional Neural Network (CNN) on edge devices. Furthermore, in this current state, the FCM codec is not agnostic to machine tasks, as different network architecture have been used for feature reduction and restoration modules of different tasks. For global adoption of FCM, and as a future direction in research, it is imperative that FCM becomes task and network agnostic, utilizing a unified feature reduction and feature restoration model capable of operating with any intermediate feature shape of any CNN and provide significant compute offload compare to edge computing.

% MPEG standards for compression aimed at machine consumption have made considerable advancements. These tools and technologies could become increasingly important as consumers demand more from their devices and electronics continue to become smarter and more interconnected. The Feature Coding for Machines MPEG standard offers significant gains, with an overall BD-Rate improvement of -93.83\% as compared to VVC only compression of intermittent features. These technologies can be incorporated into VR and AR systems, camera-equipped consumer devices, and a wide range of other applications.

\bibliographystyle{IEEEbib}
\bibliography{ref}

\vspace{12pt}
\end{document}